# Enhanced English-Sanskrit Translation using Multilingual NMT Approach

Sahinur Rahman Laskar, Anshuman Singh, Aditya Pratap Singh Sisodia

*School of Computer Science, UPES, Dehradun, India*

**Abstract.** Neural Machine Translation (NMT) is a widely accepted technique of automatic translation tasks of Natural Language Processing. However, NMT requires a sufficient amount of parallel corpora to attain a reasonable translation quality. In this work, we have considered a low-resource English-Sanskrit pair for English-to-Sanskrit and Sanskrit-to-English translation. We have performed a comparative analysis by exploring bilingual and multilingual approaches using LSTM-based Sequence-to-Sequence, and fine-tuning a pretrained model IndicTrans in a multilingual NMT approach. To investigate multilingual transfer learning, we additionally construct multilingual translation settings by incorporating Hindi along with English and Sanskrit using language-tagged data. Experimental results demonstrate that IndicTrans-based multilingual models outperform baseline results in terms of automatic evaluation metrics such as BLEU and ChrF for both directions of translation.

**Keywords:** *Multilingual NMT; English-Sanskrit; Neural Machine Translation*

## 1 Introduction

Natural Language Processing (NLP) is a field of Artificial Intelligence (AI) that focuses on enabling computers to understand and generate human language. Natural Language Processing (NLP) has seen significant advances in recent years due to the rapid development of deep learning and Neural Machine Translation (NMT) systems. Traditional statistical machine translation approaches have largely been replaced by neural architectures such as Recurrent Neural Networks (RNNs), Long Short-Term Memory (LSTM) networks [1], and, more recently, Transformer-based models. The introduction of attention mechanisms [2] and multilingual pre-trained language models has further improved translation quality across several high-resource and low-resource language pairs. In particular, multilingual NMT [3] systems have demonstrated strong transfer learning capabilities by leveraging linguistic similarities among related languages. Among classical languages, Sanskrit holds immense historical, cultural, and linguistic significance. It is one of the oldest documented Indo-Aryan languages and serves as the foundation for a large body of ancient Indian literature, philosophy, scientific texts, epics, and scriptures. Despite its importance, Sanskrit remains a low-resource language in the context of modern NLP research due to the limited availability of large-scale parallel corpora and standardized benchmarks. Furthermore, Sanskrit presents several computational challenges including rich morphology, free word order, extensive compounding, Sandhi transformations, and complex grammatical structures, making machine translation a difficult task.

In this work, we present a comparative analysis of bilingual and multilingual Neural Machine Translation systems for both Sanskrit-to-English and English-to-Sanskrit translation tasks. Our study evaluates four major configurations: (i) bilingual BiLSTM-based OpenNMT models, (ii) multilingual BiLSTM-based OpenNMT models, (iii) bilingual fine-tuned IndicTrans models, and (iv) multilingual fine-tuned IndicTrans models. To support multilingual experimentation, we construct a tagged multilingual dataset by incorporating Hindi alongside English and Sanskrit, enabling many-to-one (EN+HIN→SAN) and one-to-many (SAN→EN+HIN) translation settings. The proposed systems employ *SentencePiece*-based Byte Pair Encoding (BPE) [4], [5] and shared vocabularies to address data sparsity and improve multilingual representation learning. Experiments are conducted primarily on the *Samayik* dataset, with supplementary evaluation on the *Itihasa* corpus. Performance is evaluated using BLEU and chrF metrics under both in-domain and out-of-domain settings. Experimental results demonstrate that IndicTrans-based multilingual models achieve substantial improvements over traditional BiLSTM baselines, particularly in Sanskrit-to-English translation and out-of-domain generalization.

## 2 Related works

Sanskrit Natural Language Processing (NLP) has historically suffered from the scarcity of large-scale digitized corpora and benchmark datasets. Early efforts in Sanskrit computational linguistics primarily focused on rule-based systems, morphological analysis, dependency parsing, and Sandhi splitting due to the highly inflectional and morphologically rich nature of Sanskrit. However, the emergence of Neural Machine Translation (NMT) architectures has enabled data-driven approaches for Sanskrit translation tasks. One of the earliest large-scale Sanskrit-English translation corpora is *Itihasa*. [6]. The dataset contains more than 93,000 Sanskrit-English parallel verse pairs extracted from the Ramayana and Mahabharata. The authors demonstrated that even Transformer-based models performed poorly on the dataset, highlighting the complexity of Sanskrit translation and the limitations of existing NMT systems for morphologically rich low-resource languages. The work also emphasized the challenges introduced by Sandhi, compounding, and large vocabulary diversity in Sanskrit. More recently, Maheshwari et al. introduced *Samayik*, a contemporary English-Sanskrit parallel corpus consisting of approximately 53,000 sentence pairs collected from diverse modern domains such as spoken tutorials, educational resources, podcasts, and instructional material [7]. Unlike Itihasa, which primarily focuses on classical poetic literature, Samayik emphasizes prose-oriented contemporary Sanskrit usage. The authors additionally benchmarked several multilingual pre-trained models including mBART, IndicBART, ByT5, and IndicTrans, demonstrating improved performance on modern-domain Sanskrit translation tasks.

The development of multilingual Neural Machine Translation systems has further accelerated research in low-resource machine translation. Google's multilingual zero-shot NMT system demonstrated that a single model trained on multiple language pairs could achieve cross-lingual transfer learning and even zero-shot translation capabilities [3]. Similarly, Transformer architectures introduced in Attention Is All You Need [2] significantly improved sequence modeling and long-range dependency learning compared to recurrent architectures such as LSTMs. Recent multilingual Indic language models such as IndicTrans and IndicTrans2 have further advanced machine translation for Indian languages using large-scale Transformer encoder-decoder architectures [8]. Motivated by these developments, our work presents a comparative analysis between bilingual and multilingual NMT systems using both BiLSTM-based OpenNMT architectures and IndicTrans-based Transformer models for Sanskrit-English and English-Sanskrit translation.

## 3 System Description

This section outlines the steps involved in preparing the dataset used for experimental setup.

*A. Corpus Description*

The primary dataset used in this work is the *Samayik* corpus [7], a contemporary English-Sanskrit parallel dataset consisting of approximately 53,000 sentence pairs. Unlike earlier Sanskrit translation corpora primarily focused on classical poetic literature, *Samayik* emphasizes modern prose-oriented Sanskrit usage and contemporary linguistic expressions. The dataset was curated from multiple diverse domains including educational material, spoken tutorials, podcasts, and instructional resources, thereby providing a broader evaluation benchmark for modern Sanskrit Neural Machine Translation (NMT) systems.

The corpus is composed of five major sources: Spoken Tutorials, NIOS educational resources, Gita Sopanam, Bible translations, and Mann Ki Baat transcripts. These domains contribute to both linguistic diversity and vocabulary richness within the dataset. In particular, the Spoken Tutorials and NIOS subsets contain large amounts of instructional and technical content, while Mann Ki Baat introduces conversational and contemporary sentence structures. This diversity makes the dataset suitable for evaluating generalization capability across multiple domains.

For our experiments, the dataset is divided into training, validation, in-domain test, and out-of-domain (OOD) test sets. The final split consists of 43,493 training samples, 2,416 validation samples, and 2,417 in-domain test samples. Additionally, the Mann Ki Baat subset containing 4,047 sentence pairs is used as an out-of-domain evaluation set in order to assess the robustness and generalization capability of the translation models on previously unseen domains.

To investigate multilingual transfer learning for low resource Sanskrit translation, we further construct multilingual datasets by incorporating Hindi alongside English and Sanskrit. In this work, two multilingual configurations are explored:

EN+HIN→SAN (Many-to-one) and SAN→EN+HIN (One-to-many)

Special language tags are appended to the input sequences to distinguish target languages during multilingual training. The multilingual setting increases vocabulary diversity and enables cross-lingual transfer learning, which is particularly beneficial for Sanskrit due to the limited availability of parallel corpora.

TABLE I: Dataset Split for bilingual training

| Split | Samples | Percentage |
|---|---|---|
| Training | 43,493 | $\sim$ 83% |
| Validation | 2416 | $\sim$ 5% |
| Test | 2417 | $\sim$ 5% |
| Test OOD | 4047 | $\sim$ 8% |
| **Total** | 52,373 | |

TABLE II: Dataset Split for multilingual training

| Split | Samples | Percentage |
|---|---|---|
| Training | 86,986 | $\sim$ 90% |
| Validation | 4,832 | $\sim$ 5% |
| Test | 2417 | $\sim$ 2% |
| Test OOD | 4047 | $\sim$ 4% |
| **Total** | 96,651 | |

### *B. Experimental Setting*

Since Sanskrit is a morphologically rich and highly inflectional language, word-level tokenization often leads to severe vocabulary sparsity and rare token generation. To address this challenge, all experiments employ SentencePiece-based Byte Pair Encoding (BPE) [4], [5] with shared vocabularies across source and target languages. For the BiLSTM-based OpenNMT models, a shared vocabulary size of 32,000 subword units is used along with shared source-target embeddings to improve multilingual representation learning and reduce parameter redundancy. The first set of experiments utilizes the OpenNMT framework [9] with a sequence-to-sequence encoder-decoder architecture based on Bidirectional Long Short-Term Memory (BiLSTM) networks. The encoder consists of a two-layer bidirectional LSTM network, while the decoder employs a two-layer LSTM architecture with MLP-based global attention. Both embedding dimension and hidden state size are set to 512. Dropout regularization of 0.3 and label smoothing of 0.1 are applied during training to improve model generalization. The OpenNMT systems are trained under both bilingual and multilingual settings for Sanskrit-to-English and English-to-Sanskrit translation tasks.

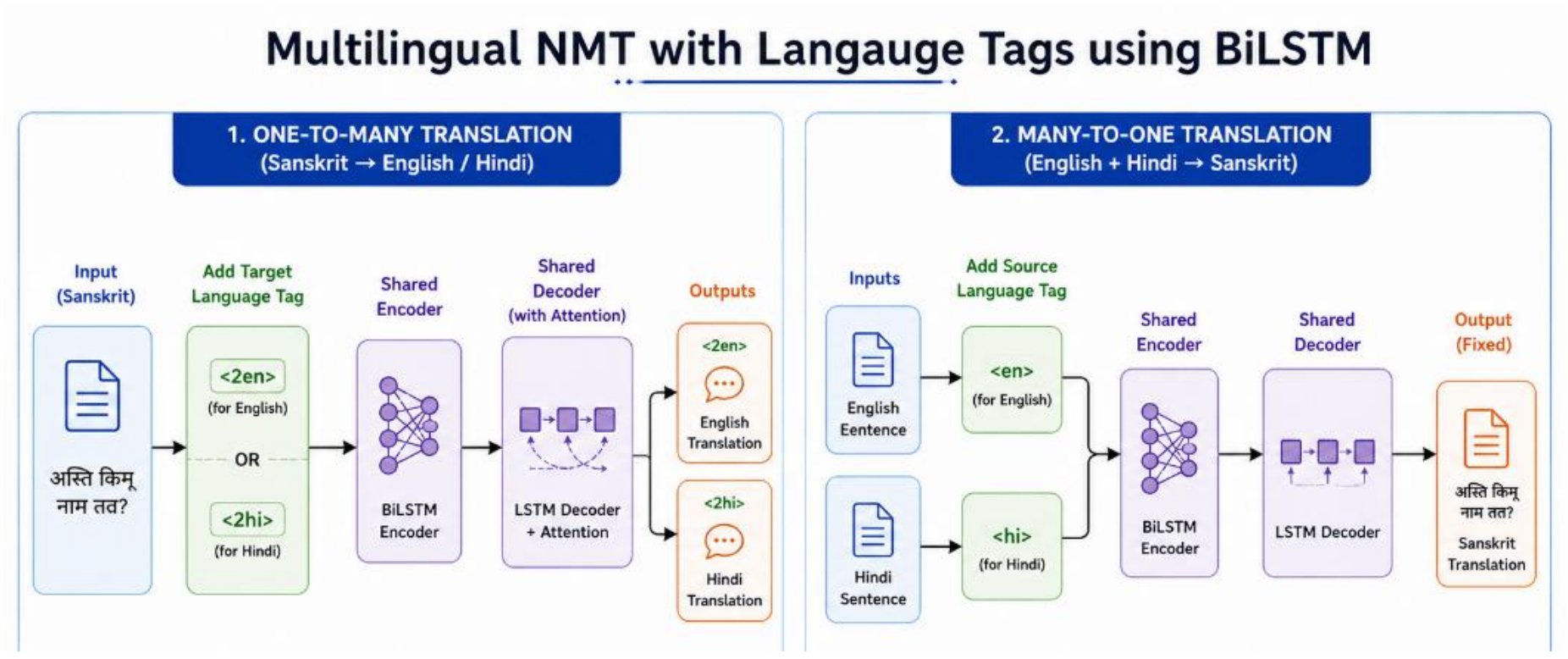

Fig. 1: Multilingual NMT with Language Tags using BiLSTM

The second set of experiments is based on IndicTrans2- 1B. IndicTrans2 employs 18 encoder layers and 18 decoder layers with approximately one billion parameters. The model uses multilingual SentencePiece tokenization along with language control tags to support multilingual translation. Fine-tuning is performed separately for bilingual and multilingual datasets in both translation directions. Training for the OpenNMT systems is performed using the Adam optimizer with token-based batching and early stopping based on validation perplexity. For IndicTrans2 fine-tuning, FP16 mixed precision training is employed on NVIDIA A100 GPUs in order to efficiently handle the computational requirements of the 1B parameter model. Evaluation is conducted using both BLEU [10] and chrF [11] metrics under in-domain and out-of-domain settings. While BLEU evaluates word-level n-gram precision, chrF provides a character-level evaluation metric better suited for morphologically rich languages such as Sanskrit.

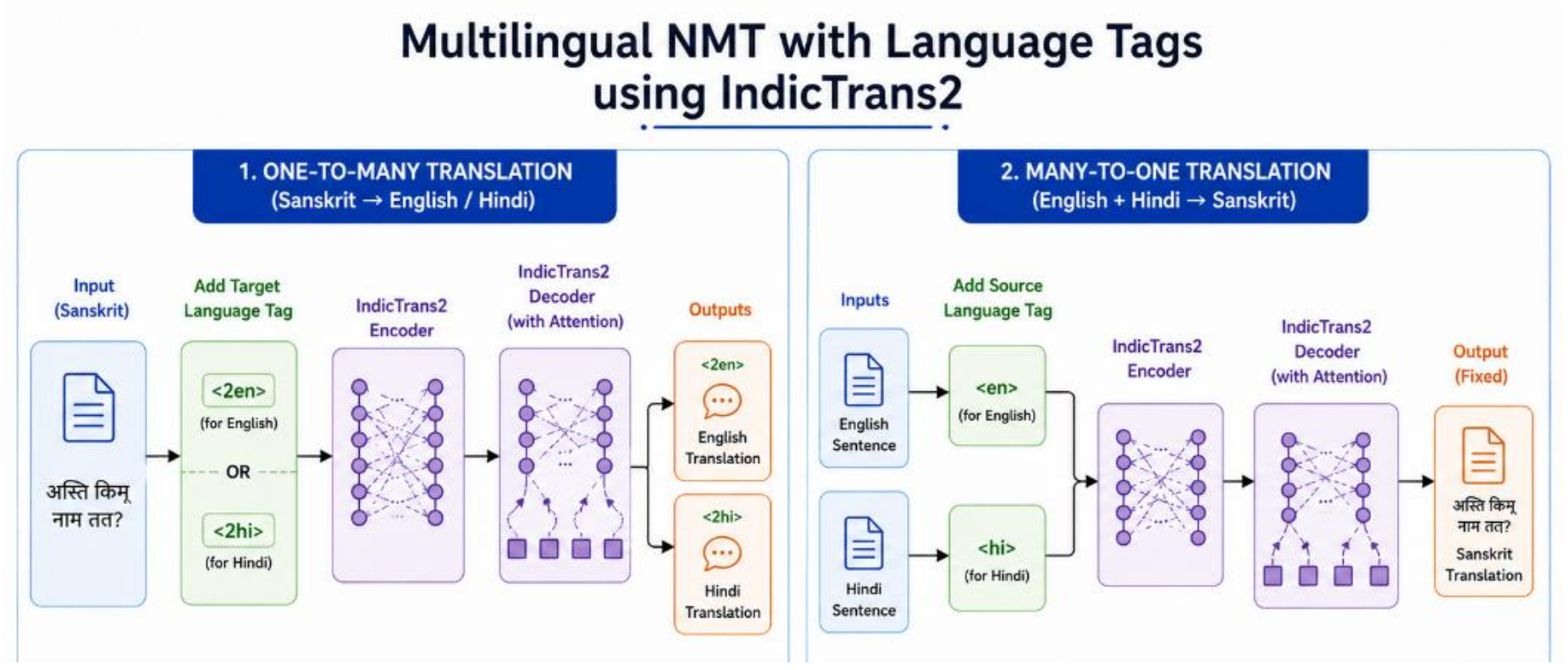

Fig. 2: Multilingual NMT with Language Tags using IndicTrans2

## 4 Results and analysis

In this section, we evaluate the performance of bilingual and multilingual NMT systems for both Sanskrit-to-English (Sa-En) and English-to-Sanskrit (En-Sa) translation tasks. The models are evaluated under both in domain and out-of-domain (OOD) settings using BLEU and chrF metrics. Experiments are conducted primarily on the *Samayik* dataset, with supplementary evaluation on the *Itihasa* corpus.

*A. Sanskrit to English Translation*

Tables III and V present the Sanskrit-to-English trans lation results on the *Samayik* and *Itihasa* datasets respectively. On the *Samayik* dataset, IndicTrans-based systems significantly outperform the BiLSTM-based OpenNMT models across both in-domain and OOD settings. The multilingual IndicTrans model achieves the best performance among our systems with an in-domain BLEU score of 30.49 and an OOD BLEU score of 16.42. A similar trend is observed on the Itihasa dataset, where IndicTrans models again outperform recurrent BiLSTM architectures. The multilingual IndicTrans model achieves the best results with BLEU scores of 8.66 (in-domain) and 8.43 (OOD), while also obtaining the highest chrF scores. In contrast, the BiLSTM systems show substantially lower performance and weaker generalization capability on unseen domains. The higher chrF scores achieved by IndicTrans models across both datasets indicate better handling of Sanskrit morphology and subword-level representations. Overall, multilingual training improves Transformer-based systems more effectively than recurrent sequence-to-sequence architectures.

*B. English to Sanskrit Translation*

Tables IV and VI present the English-to-Sanskrit translation results on the Samayik and Itihasa datasets respectively. Compared to Sanskrit-to-English translation, all systems achieve considerably lower BLEU scores, indicating the increased complexity of Sanskrit text generation. On the Samayik dataset, IndicTrans-based systems consistently outperform the BiLSTM models in both bilingual and multilingual configurations. The multilingual Indic Trans model achieves the best in-domain performance among our systems with a BLEU score of 10.05 and a chrF score of 39.50. On the Itihasa dataset, performance further decreases across all systems due to the complexity of classical Sanskrit verse translation. Nevertheless, IndicTrans models continue to achieve comparatively stronger chrF scores and better semantic consistency than BiLSTM architectures. The multilingual IndicTrans model achieves the best overall chrF performance with scores of 27.13 (in-domain) and 19.48 (OOD), while also matching the highest OOD BLEU score of 0.25. The comparatively lower performance in this translation direction can be attributed to the rich morphology, free word order, Sandhi transformations, and complex grammatical structure of Sanskrit, which makes accurate target-side generation substantially more challenging.

*C. Comparative Analysis*

The experimental results demonstrate that Transformer-based IndicTrans models consistently outperform BiLSTM-based OpenNMT systems across both datasets and translation directions. Multilingual training further improves the performance of IndicTrans, particularly for Sanskrit-to-English OOD translation tasks. The experiments also reveal that English-to-Sanskrit translation remains considerably more difficult than Sanskrit-to-English translation due to the complexity of Sanskrit generation. Furthermore, multilingual transfer learning using Hindi improves translation quality and generalization capability, especially for Transformer-based multilingual systems. Tables VII and VIII present qualitative examples from the out-of-domain test set, highlighting the superior flu ency and semantic consistency of IndicTrans-based multilingual models compared to recurrent BiLSTM systems.

TABLE III: Sanskrit to English Results

| System | In-domain (2,417) | | OOD (4,047) | |
|---|---|---|---|---|
| | **BLEU** | **chrF** | **BLEU** | **chrF** |
| *Samayik Baseline* | | | | |
| ByT5 | 31.10 | 55.70 | 5.40 | 29.00 |
| IndicBART | 29.79 | 50.14 | 5.30 | 27.70 |
| Indictrans | 37.00 | 58.20 | 13.10 | 37.50 |
| *Our Models* | | | | |
| BiLSTM | 21.50 | 36.50 | 1.80 | 16.92 |
| BiLSTM (Multilingual) | 18.12 | 33.18 | 1.51 | 16.92 |
| IndicTrans (Fine-tuned) | 26.06 | 53.15 | 15.16 | 43.56 |
| **IndicTrans (Multilingual)** | **30.49** | **53.34** | **16.42** | **42.50** |

TABLE IV: English to Sanskrit Results

| System | In-domain (2,417) BLEU | In-domain (2,417) chrF | OOD (4,047) BLEU | OOD (4,047) chrF |
|---|---|---|---|---|
| *Samayik Baseline* | | | | |
| ByT5 | 28.70 | 44.40 | 7.00 | 21.40 |
| IndicBART | 25.45 | 43.47 | 6.90 | 22.40 |
| Indictrans | 11.30 | 46.60 | 0.60 | 26.70 |
| *Our Models* | | | | |
| BiLSTM | 7.82 | 21.81 | 0.078 | 13.65 |
| BiLSTM (Multilingual) | 7.81 | 21.68 | 0.096 | 13.75 |
| IndicTrans (Fine-tuned) | 9.67 | 39.09 | 1.11 | 27.52 |
| **IndicTrans (Multilingual)** | **10.05** | **39.50** | **1.01** | **26.97** |

TABLE V: Itihasa: Sanskrit to English Results

| System | In-domain (11,721) BLEU | In-domain (11,721) chrF | OOD (4047) BLEU | OOD (4047) chrF |
|---|---|---|---|---|
| BiLSTM | 6.51 | 27.86 | 0.28 | 18.17 |
| BiLSTM (Multilingual) | 4.12 | 20.56 | 0.17 | 15.69 |
| IndicTrans (Fine-tuned) | 7.84 | 31.22 | 13.80 | 40.67 |
| **IndicTrans (Multilingual)** | **8.66** | **30.12** | **8.43** | **29.76** |

TABLE VI: Itihasa: English to Sanskrit Results

| System | In-domain (11,721) BLEU | In-domain (11,721) chrF | OOD (4,047) BLEU | OOD (4,047) chrF |
|---|---|---|---|---|
| BiLSTM | 1.205 | 24.809 | 0.016 | 12.81 |
| BiLSTM (Multilingual) | 0.64 | 18.42 | 0.014 | 8.49 |
| IndicTrans (Fine-tuned) | 0.57 | 25.79 | 0.25 | 19.31 |
| **IndicTrans (Multilingual)** | **0.90** | **27.13** | **0.25** | **19.48** |

TABLE VII: Qualitative Example: English → Sanskrit (OOD)

| **System** | **Translation** |
|---|---|
| **Source** | The number of suggestions and phone calls for this episode is many times the usual. |
| Google Translate | अस्य प्रकरणस्य सुझावस्य, दूरभाषस्य च संख्या सामान्यतः बहुगुणा अस्ति । |
| BiLSTM | अमरकोशस्य प्रथमांशस्य प्रथमांशस्य प्रथमांशस्य च एकत्रीकरणं भवति। |
| BiLSTM (Multilingual) | सामान्यक्षराणि अपि च अक्षरसङ्ख्याः, अपि च ▯▯▯' इति आदेशान् उपयुज्यन्ते। |
| IndicTrans (FT) | अस्य प्रकरणस्य कृते सूचनानां दूरभाषाणां च सङ्ख्या सामान्यस्य अपेक्षया बहुगुणिता अस्ति। |
| IndicTrans (Multilingual) | अस्य प्रकरणस्य कृते सूचनानां दूरभाषाणां च सङ्ख्या सामान्यस्य अपेक्षया बहुवारम् अस्ति। |
| **Reference** | अस्य मनोगतस्य कृते सम्प्राप्यमाणा परामर्श-दूरभाषाकारणानां संख्या सामान्यतः अनेकधा अधिकास्ति। |

TABLE VIII: Qualitative Example: Sanskrit → English (OOD)

| **System** | **Translation** |
|---|---|
| **Source** | अद्य जगति दक्ष-जनानां कृते अवसराणां न्यूनता नास्ति । |
| Google Translate | There is no shortage of opportunities for skilled people in the world today. |
| BiLSTM | However, I don't have a list of people at the end of the world. |
| BiLSTM (Multilingual) | Now, I don't want to be less than the world. |
| IndicTrans (FT) | There is no dearth of opportunities for skilled people in the world today. |
| IndicTrans (Multilingual) | There is no dearth of opportunities for skilled people in the world today. |
| **Reference** | There is no dearth of opportunities for skilled people in the world today. |

## 5 Conclusions

In this work, we presented a comparative study of bilingual and multilingual Neural Machine Translation systems for Sanskrit-to-English and English-to-Sanskrit translation tasks. The study evaluated BiLSTM-based OpenNMT architectures alongside Transformer-based IndicTrans models using the Samayik dataset [7] under both in-domain and out-of-domain evaluation settings. Experimental results demonstrate that IndicTrans-based models consistently outperform traditional BiLSTM sequence-to-sequence systems across both translation directions. In particular, multilingual IndicTrans models achieved the best overall performance, highlighting the effectiveness of multilingual transfer learning and pretrained Transformer architectures for low-resource Sanskrit translation. The experiments also revealed that English-to-Sanskrit translation remains significantly more challenging than Sanskrit-to-English translation due to the rich morphology, free word order, and complex grammatical structure of Sanskrit. Furthermore, multilingual training using Hindi as an auxiliary language improved generalization capability, especially for out-of-domain Sanskrit-to-English translation. Overall, this work highlights the advantages of multilingual Transformer-based approaches for low-resource Indic language translation and establishes stronger comparative baselines for future Sanskrit Neural Machine Translation research. .